# Video retrieval based on deep convolutional neural network

Yajiao Dong
School of Information and Electronics,
Beijing Institution of Technology, Beijing, China
yajiaodong@bit.edu.cn

Jianguo Li
School of Information and Electronics,
Beijing Institution of Technology, Beijing, China
jianguoli@bit.edu.cn

## ABSTRACT
Recently, with the enormous growth of online videos, fast video retrieval research has received increasing attention. As an extension of image hashing techniques, traditional video hashing methods mainly depend on hand-crafted features and transform the real-valued features into binary hash codes. As videos provide far more diverse and complex visual information than images, extracting features from videos is much more challenging than that from images. Therefore, high-level semantic features to represent videos are needed rather than low-level hand-crafted methods. In this paper, a deep convolutional neural network is proposed to extract high-level semantic features and a binary hash function is then integrated into this framework to achieve an end-to-end optimization. Particularly, our approach also combines triplet loss function which preserves the relative similarity and difference of videos and classification loss function as the optimization objective. Experiments have been performed on two public datasets and the results demonstrate the superiority of our proposed method compared with other state-of-the-art video retrieval methods.

## 1. INTRODUCTION
With the development of technology, video is everywhere on the internet. And video retrieval methods have become an important challenge nowadays, which aim to find those most relevant videos from a dataset for a query video in an efficient and accurate manner. In contrast to images, videos provide various and complex visual patterns consisting of low-level visual content in each frame as well as high-level semantic content across frames, which makes video retrieval more challenging than image retrieval.

The key points of video retrieval lie in the technology of video feature extraction and video feature similarity measurement. Traditional video retrieval methods generally employ hand-crafted methods to extract features and mainly focus on how to use the video features to achieve the optimal binary coding, namely, finding an appropriate hash mapping function. Conventional hashing algorithms learn binary hashing codes whose distance is correlated to the similarity relationship of the original input data [1, 2, 3]. Locality sensitive hashing (LSH) [1] adopts random projections to map original data into a low-dimensional feature space, and then transforms real-valued features into binary codes. Semantic hashing (SH) [2] employs a multi-layers Restricted Boltzmann Machines (RBM) to learn compact binary codes for input data. Iterative quantization (ITQ) [3] applies iterative optimization strategy to find projections with minimal binarization loss. Even though these methods have been proved to be relatively effective, the binary hash codes still cannot represent the input data accurately. These traditional binary hash coding methods all employ hand-crafted features to compare original data similarity, which is not effective due to the differences between the high-level semantical similarity that human can observe and the low-level visual similarity that machines can learn.

With the rapid development of deep learning, more methods based on deep learning have provided us with the prospect of efficient and accurate video retrieval. Deep learning can acquire high-level semantical features by combining lower-level visual features and Deep Convolutional Neural Networks (CNNs) [4] have proved to be versatile image representation tools with strong generalization capabilities, which makes CNNs an indispensable part of efficient video retrieval methods. Recently, inspired by the success of deep learning in image recognition, several video retrieval methods have incorporated hash functions into deep learning architectures [5][6]. For example, [5] applies three-layer hierarchical neural networks to learn discriminative projection matrix, assuming the video pairwise information is available. However, their method is unable to take the advantage of deep transfer learning, thus makes the binary codes less effective. Besides, [6] proposes a deep encoder-decoder framework, where a two-layer Long Short Term Memory (LSTM) unit followed by a binarization layer can directly encode the video features into binary hashing codes. But the objective function based on minimizing reconstruction error does not seem to preserve the neighborhood structure of data, which is important for a similarity retrieval task.

Rapid video retrieval based on deep neural networks is receiving more attention. In this paper, an end-to-end supervised and hashing framework is proposed, in which feature extraction and hash mapping function are integrated in a network to achieve an end-to-end optimization.

## 2. PROPOSED APPROACH
### 2.1 Architecture
In this paper, we propose an architecture of deep convolution neural networks designed for video retrieval, as shown in Figure 1. In the training phase, this architecture accepts input videos in a triplet form. Three sets of video frames are selected randomly from the three input videos and each set has the same number of video frames. Given three sets of video frames, the pipeline of the proposed architecture contains three parts: 1) The unified feature representation of each input frame is extracted from the deep convolutional neural network, and then video frame features of each set are fused into video features by weighted average in order to simplify the complexity of the network; 2) The first fully connected layer is followed by a sigmoid layer designed to learn similarity-preserving binary-like codes, and the second fully connected layer has $k$ nodes which are equals to the number of categories; 3) The final part is loss function combined by classification loss and triplet loss. In the retrieving phase, binary hashing codes are generated by binarization which maps the binary-like outputs into 0 or 1. Then exclusive-or operation is conducted on the binary codes of the query video and the binary codes of the videos stored in datasets to obtain the corresponding Hamming distance and find the video with the highest similarity.

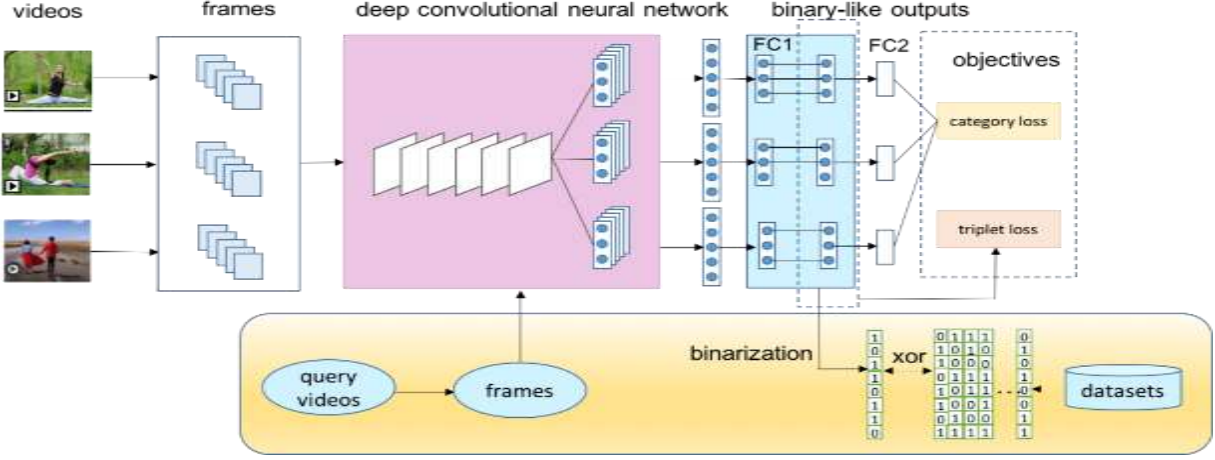

**Figure 1. Overview of the proposed network architecture including the training network and the retrieving network.**

## 2.2 Loss Function

In the proposed deep network architecture, we present a variant of the triplet loss in [7] to preserve the relative similarity and difference of videos. Specifically, given the triplet of training videos in the form of $(X, X+, X-)$ in which $X$ is more similar to $X+$ than to $X-$, the goal is to find an appropriate mapping $F(.)$ so that the binary-like code $F(X)$ is closer to $F(X+)$ than to $F(X-)$. Accordingly, the triplet loss is defined by

$$l_1 = \max(\|F(X) - F(X^+)\|_2^2 - \|F(X) - F(X^-)\|_2^2 + m, 0) \quad (1)$$

where $\|\cdot\|_2$ is the L2-norm, denoting the distance between two vectors, and $m > 0$ is a margin threshold parameter. If there is not the parameter $m$ in the triplet loss function, minimizing the loss function will result in the representation of each video tending to be zero, seriously affecting the system performance.

Classification loss is defined by

$$l_2 = -\sum_{n=1}^{3N}\sum_{k=1}^{K} t_{nk} \ln p_{nk} \quad (2)$$

where $N$ is the number of triplets of training videos, $K$ is the number of categories in one dataset, $\boldsymbol{t_n}$ is a binary vector with element $t_{nk}$ and the number of elements in $\boldsymbol{t_n}$ equals $K$. If the input video belongs to class $k$, all elements of $\boldsymbol{t_n}$ are zeros except for element $k$, which equals to one. $p_{nk}$ is the possibility of the nth input belonging to the kth class. It is noticed that eq. (2) is known as the cross-entropy loss function for the multiclass classification problem.

By combining triplet loss function and classification loss function together, we obtain the overall loss function as follows:

$$l = \alpha \cdot \sum_{n=1}^{N} l_1 + \beta \cdot l_2 \quad (3)$$

Where $\alpha$ and $\beta$ are hyper-parameters that balance the two loss functions. Finally, we update the parameters of the network by minimizing the overall loss function.

## 3. EXPERIMENTS

To evaluate the effectiveness of our proposed method for scalable video retrieval, we conducted experiments on two public video datasets named the UCF101[8] datasets and HMDB51 [9] datasets. The details of the experimental results are described in the following sections.

### 3.1 Datasets

We empirically evaluate the proposed architecture on the UCF-101 [8]and HMDB-51 [9] datasets. UCF101 consists of 13320 realistic action videos collected from YouTube in the datasets, having 101 action categories. It covers a broad set of activities such as sports, musical instruments, and human-object interaction. HMDB-51 contains about 7,000 videos collected from a variety of sources ranging from digitized movies to YouTube and has been classified into 51 distinct action categories, each containing at least 101 videos.

### 3.2 Results on Video Retrieval

To evaluate the effectiveness of our proposed architecture, we compare our method with several traditional hashing methods, including LSH [1], ITQ [3], SGH [10], Spectral hashing (SH) [2], AGH [11], and Deep Hashing (DH) [5], PCA-RR [12], SELVE[13] on the UCF101 dataset and HMDB51 dataset. Among these eight approaches, only Deep Hashing takes advantage of deep neural networks for learning binary hashing codes with an end-to-end architecture as our method. To be fair, the inputs of other methods are also video features extracted from the same deep neural network as our method rather than hand-crafted features. However, the learning of hashing function is separated from the feature extraction in traditional methods.

Without loss of generality, we compare the performance of these methods with respect to different code lengths. Table 1 shows the video retrieval results based on the mean Average Precision (mAP) of the top 10 on UCF101 dataset. Our proposed approach has obvious improvement compared with traditional best retrieval performance by 4.3%, 4.6%, 4.8% and 4.6% mAP with respect to 512, 256, 128 and 64 binary hash bits respectively. Even though compared with Deep Hashing, our method also has performance improvement to some extent, which proves that the triplet loss function proposed in this paper is reasonable and effective. Furthermore, according to the results, we find that the longer the hash bits are, the better performance our approach can achieve. In addition, we also conducted experiments on the HMDB51 dataset based on the mean Average Precision (mAP) of the top 20. The experimental results are shown in Figure 2, which further proves the effectiveness of our method.

**Table 1. Performance comparison of different video retrieval algorithms on the UCF101 dataset. This table shows the mean Average Precision (mAP) of top10.**

| method | 512bits | 256bits | 128bits | 64bits |
| --- | --- | --- | --- | --- |
| AGH | 0.449 | 0.491 | **0.515** | 0.495 |
| PCA-RR | **0.724** | 0.717 | 0.689 | 0.650 |
| LSH | **0.736** | 0.710 | 0.671 | 0.605 |
| SH | 0.616 | 0.641 | **0.644** | 0.619 |
| ITQ | **0.757** | 0.750 | 0.735 | 0.701 |
| SGH | **0.697** | 0.685 | 0.491 | 0.323 |
| SELVE | **0.683** | 0.665 | 0.683 | 0.660 |
| DH | **0.790** | 0.778 | 0.759 | 0.723 |
| **ours** | **0.800** | **0.796** | **0.783** | **0.747** |

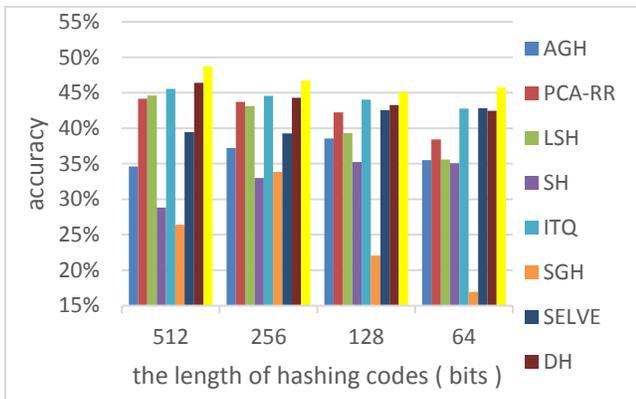

Figure 2. Comparison of video retrieval performance of our method and other hashing methods on HMDB51 dataset. This figure shows the mAP of top20.

## 4. CONCLUSION

In this paper, we have presented a supervised deep learning framework to achieve video retrieval. We employ deep convolutional neural networks to extract high-level semantic features of input videos and map real-valued representations into binary hash codes in order to simplify the complexity of computation. In our approach, two optimization loss functions are proposed and we update the network parameters through minimizing the two loss functions together. Experimental results on two video datasets clearly demonstrate that our approach achieves better performance than other video retrieval methods.